\documentclass[10pt,twocolumn,letterpaper]{article}

\usepackage{wacv}
\usepackage{times}
\usepackage{epsfig}
\usepackage{graphicx}
\usepackage{amsmath}
\usepackage{amssymb}

\usepackage{color}
\usepackage{subcaption}
\usepackage{algorithm,algpseudocode}
\usepackage{bbm}
\usepackage{tablefootnote}
\usepackage{adjustbox}
\usepackage[font=small]{caption}

\def\wacvPaperID{1281} 

\wacvfinalcopy 

\ifwacvfinal
\def\assignedStartPage{9876} 
\fi

\ifwacvfinal
\usepackage[breaklinks=true,bookmarks=false]{hyperref}
\else
\usepackage[pagebackref=true,breaklinks=true,colorlinks,bookmarks=false]{hyperref}
\fi

\ifwacvfinal
\else
\fi

\begin{document}
\pagenumbering{gobble}
\title{DACS: Domain Adaptation via Cross-domain Mixed Sampling}

\author{Wilhelm Tranheden$^{1,2}$\thanks{Equal contribution.},
Viktor Olsson$^{1,2}${\footnotemark[\value{footnote}]},
Juliano Pinto$^{1}$,
Lennart Svensson$^{1}$\\
$^1$Chalmers University of Technology, Gothenburg, Sweden\\
$^2$Volvo Cars, Gothenburg, Sweden\\
{\tt\small \{wilhelm.tranheden,viktor.olsson.3\}@volvocars.com, \{juliano,lennart.svensson\}@chalmers.se}
}

\maketitle

\begin{abstract}
   Semantic segmentation models based on convolutional neural networks have recently displayed remarkable performance for a multitude of applications. However, these models typically do not generalize well when applied on new domains, especially when going from synthetic to real data. In this paper we address the problem of unsupervised domain adaptation (UDA), which attempts to train on labelled data from one domain (source domain), and simultaneously learn from unlabelled data in the domain of interest (target domain). Existing methods have seen success by training on pseudo-labels for these unlabelled images. Multiple techniques have been proposed to mitigate low-quality pseudo-labels arising from the domain shift, with varying degrees of success. We propose DACS: Domain Adaptation via Cross-domain mixed Sampling, which mixes images from the two domains along with the corresponding labels and pseudo-labels. These mixed samples are then trained on, in addition to the labelled data itself. We demonstrate the effectiveness of our solution by achieving state-of-the-art results for GTA5 to Cityscapes, a common synthetic-to-real semantic segmentation benchmark for UDA.
\end{abstract}


\section{Introduction}
Deep neural networks have significantly advanced the state of the art for the task of semantic segmentation \cite{long2014fully,DeepLabv1,PSPNet}, displaying remarkable generalization abilities. Results have in large been presented for datasets where training and test domains are similar or identical in distribution. In real-world scenarios, however, a domain shift may occur, where the training data (source domain) is significantly different to the data encountered in inference (target domain) for the intended application. A common practice when dealing with domain shift is to annotate some data from the domain of interest and re-train (fine-tune) the network on this new data. Additionally, Semi-Supervised Learning (SSL) methods for semantic segmentation have been proposed \cite{Hung,Mittal,French,Feng,chen2020leveraging,ClassMix}, effectively training on a small amount of labelled data by relying on complementary learning from unlabelled data. These approaches are not always feasible as sometimes no annotations at all are accessible in the target domain. 

Unsupervised Domain Adaptation (UDA) deals with the problem where labelled data is available for a source domain, but only unlabelled data is available for the target domain. The field has generated a lot of interest due to its potential for effectively utilizing synthetic data in training deep neural networks. Semantic segmentation in particular has been explored in recent research \cite{tsai2018learning,luo2019significanceaware,luo2018taking,yang2019adversarial,tsai2019domain,vu2018advent,zhang2018fully,zou2018domain,zou2019confidence,li2019bidirectional,yang2020contextaware,zheng2019unsupervised,zheng2020rectifying} due to its potential benefits in applications such as autonomous driving, where the associated annotation cost is high and we may prefer to generate cheap synthetic data.


A technique originally proposed for SSL is pseudo-labelling \cite{pseudo-label} (or self-training); training on artificial targets based on the class predictions of the network. Pseudo-labelling was later adapted to UDA \cite{zou2018domain,zou2019confidence,zheng2020rectifying}, where certain modifications were introduced to compensate for the domain shift. One of the earlier works on pseudo-labelling for UDA \cite{zou2018domain} pointed out that pseudo-labelling naively applied to UDA tends to bias the predictions of the network to easy-to-predict classes, causing difficult classes to stop being predicted during training. To combat this collapse, they propose class-balanced sampling of the pseudo-labels. Additional difficulties of erroneous pseudo-labels, owing to the domain shift, have prompted later research to add modules for uncertainty estimation \cite{zou2019confidence,zheng2020rectifying}. We note that in existing methods for correcting erroneous pseudo-labels, certain images in the target domain are over-sampled, and low confidence pixels within images filtered out. Many pixels of low confidence are aligned with predictions at semantic boundaries \cite{li2017pixels,zhu2018improving,ClassMix}, thus leading to a diminished training signal there. Circumventing these issues offers an opportunity to better leverage available data in the target domain.
In this paper we propose Domain Adaptation via Cross-domain mixed Sampling, or DACS for short. The key idea here is that we form new, augmented samples by mixing pairs of images from different domains. In particular, we use the ground-truth semantic map to extract a few classes from a source domain image and paste them
onto an unlabelled target domain image. To construct pseudo-labels for the new image, we mix the source domain labels with pseudo-labels created from the target domain image.
Mixing across domains this way leads to parts of the pseudo-labels being replaced by parts of the ground-truth semantic maps, ensuring that over the course of training all classes will border pixels from the other domain.

Our method is inspired by recently proposed methods of using mixed samples for SSL \cite{French,StructuredLoss,ClassMix}, and we show that applying the naive implementation of these methods directly to UDA causes classes to become conflated, i.e., certain classes are confused with others, similarly to the previously mentioned findings in \cite{zou2018domain} for naive pseudo-labelling. This problem is effectively solved by the cross-domain mixing in DACS. In contrast to existing methods for correcting erroneous pseudo-labels, we do not get rid of training data from over-sampling by class or confidence, and are able to efficiently learn from the entire unlabelled target domain dataset during the course of training. We demonstrate the effectiveness of our method by applying it on two synthetic-to-real unsupervised domain adaptation benchmarks, GTA5 $\rightarrow$ Cityscapes and SYNTHIA $\rightarrow$ Cityscapes.


In summary, our main contributions are: (1) We introduce an algorithm that mixes images from the target domain with images from the source domain, creating new, highly perturbed samples and use these for training. (2) We show that mixing of images across domains to a high degree solves the problem of conflating classes.
(3) We present improvements over the state of the art in UDA for the GTA5 $\rightarrow$ Cityscapes benchmark.

\section{Related Work}
For semantic segmentation, predominant methods for UDA include adversarial-based \cite{tsai2019domain}, self-training \cite{zou2018domain,zou2019confidence,zheng2020rectifying} and consistency-based approaches \cite{chen2020crdoco,zhou2020uncertaintyaware}. This section reviews these related UDA methods in more detail, as well as some of their counterparts in SSL, as the two tasks are closely related. For a more extensive review of UDA in semantic segmentation, see \cite{toldo2020unsupervised}.

\vspace{0.4cm}
\noindent{\bf Domain alignment.} A lot of existing research in UDA has focused on various aspects of adversarial learning in order to bridge the gap existing between source and target domains, minimizing differences between the distributions. This can be targeted at different levels such as at the pixel level \cite{bousmalis2016unsupervised,hoffman2017cycada,wu2019ace, yang2020labelddriven}, feature map level \cite{yue2019domain,luo2019significanceaware,zhang2019manifold,hong2018conditional,zhu2018penalizing,wang2020differential,DBLP:conf/nips/ZhangZ0T19} or semantic level \cite{tsai2019domain,Wang2019classspecificrecon}. In particular, alignment on the semantic level has been explored with similar methods for both UDA \cite{tsai2019domain} and SSL \cite{Hung}. The key idea is viewing the segmentation network as a generator in a generative adversarial network setup, to encourage realistic semantic maps to be predicted. This approach is viable for UDA because even though the source and target domain images are quite different, it is often reasonable to assume that the corresponding semantic maps are similar in terms of spatial layout and local context. Our proposed method should also benefit from this similarity in output space, since mixing images across domains will then lead to semantic classes being placed in contexts with more similar semantics.

Other approaches to aligning the source and target domains include channel-wise alignment of feature maps \cite{wu2018dcan}, done differently for "stuff" classes and "things" classes in \cite{wang2020differential}, transfer of texture from target to source \cite{DBLP:conf/cvpr/KimB20a}, and transfer of the low-frequency part of the spectrum in Fourier space from the target domain to the source domain \cite{yang2020fda}.

\vspace{0.4cm}
\noindent {\bf Pseudo-labelling.} In contrast, methods based on pseudo-labelling \cite{pseudo-label}, or self-training, directly train on the target domain data by the use of pseudo-labels, which are artificial targets for training, created from class predictions of the network. However, UDA problems characteristically suffer from large domain gaps, i.e., considerable differences in data distributions, which give rise to faulty pseudo-labels. One of the problems is a bias in the target domain towards initially easy-to-transfer classes \cite{zou2018domain,zou2019confidence}, where some classes are merged, meaning certain classes are never predicted. A similar phenomenon has been observed in UDA for classification \cite{wu2020entropy}, where trivial predictions for the target domain were identified to occur when applying entropy regularization. For semantic segmentation, entropy regularization methods have observed a bias towards easy-to-transfer classes \cite{vu2018advent} as well. 

The close connection between entropy regularization and pseudo-labelling was pointed out in the original proposal of pseudo-labelling \cite{pseudo-label}. As pseudo-labelling and entropy regularization strive to minimize entropy in the predictions of unlabelled data, and since predictions with conflated, i.e., merged, classes have lower entropy, entropy minimization is a reasonable explanation for why conflation of classes can occur for large enough domain shifts. To combat the problem of faulty pseudo-labels for UDA in semantic segmentation, existing works have suggested careful selection and adjustment procedures, accounting for the domain gap. Variants include specialised sampling \cite{zou2018domain, DBLP:journals/corr/abs-2008-12197} and handling of uncertainty \cite{zou2019confidence,zheng2020rectifying}. 

Our proposed solution also makes use of pseudo-labelling, while cross-domain mixing of samples offers a simple solution to the class conflation problem by injecting reliable entropy from the source domain ground-truth labels into the pseudo-labels of the target domain. 
Even though we make use of pseudo-labels generated from unperturbed images from the target domain, they are only used for training after images and labels are mixed across domains. Therefore, we categorize our approach along another line of methods, namely consistency regularization, which is consistent with related SSL methods.

\vspace{0.4cm}
\noindent{\bf Consistency Regularization.} The key idea behind consistency regularization is that predictions on unlabelled samples should be invariant to perturbations. In SSL the perturbations have typically been based on image augmentations \cite{French,ClassMix,berthelot2019mixmatch,xie2019unsupervised,FixMatch,StructuredLoss}. For UDA this technique has instead been used to complement minimization of distribution discrepancies on an image level, with consistency based around image-to-image translations \cite{chen2020crdoco, zhou2020uncertaintyaware}.

\vspace{0.4cm}
\noindent{\bf Mixing.}
The kind of augmentation technique known as mixing has successfully been used for both classification and semantic segmentation \cite{MixUpOriginal,CutMix,berthelot2019mixmatch,French,MilkingCowMask,StructuredLoss,ClassMix,tokozume2018between}. The idea is to combine pixels from two training images, thereby creating a new, highly perturbed, sample. This can be done by interpolating between the pixel values of the two images, as suggested by Zhang et al. in their mixup algorithm \cite{MixUpOriginal}. It can also be done by having one set of pixels coming from one image and another set coming from a second image. 
In the latter method, the selection of pixels can be quantified by a binary mask, where the mask is one for the pixels selected from the first image and zero for pixels selected from the second image.
For the case of semantic segmentation, the same mixing is also performed on the segmentation maps.

Mixing has proven to be successful in SSL for semantic segmentation, where the strong augmentations are used on unlabelled images for consistency regularization. Kim et al. \cite{StructuredLoss} and French et al. \cite{French} use the binary mixing method CutMix, proposed by Yun et al. \cite{CutMix}, where a rectangular region is cut from one image and pasted on top of another. This technique was further developed in the ClassMix algorithm \cite{ClassMix}, where the mask used for mixing is instead created dynamically based on the predictions of the network. Specifically, the algorithm selects half of the classes predicted for a given image, and the corresponding pixels are pasted onto a second image, forming a strongly perturbed image, in which the semantic borders of objects are still being followed to a high degree.

Our proposed method, DACS, adapts this technique to mix images across domains. This differs from existing consistency-based approaches for UDA in that we do not attempt to combine consistency regularization with alignment of image distributions, instead, we enforce consistency between predictions of images in the target domain and images mixed across domains.


\section{Method}
\label{sec:method}

This section details our proposed approach for unsupervised domain adaptation: Domain Adaptation via Cross-domain mixed Sampling, or DACS for short. We start in Section \ref{sec:naivemix} by discussing the pitfalls of applying mixing-based consistency regularization as used for SSL directly to UDA, that is, mixing images only within the unlabelled dataset. In Section \ref{sec:DACS} we then describe our solution DACS and explain how it solves these problems.
We then conclude with a description of the loss function and the training procedure used. 


\subsection{Naive Mixing to UDA}
\label{sec:naivemix}
\begin{figure}[t]
    \centering
    \includegraphics[width=\columnwidth]{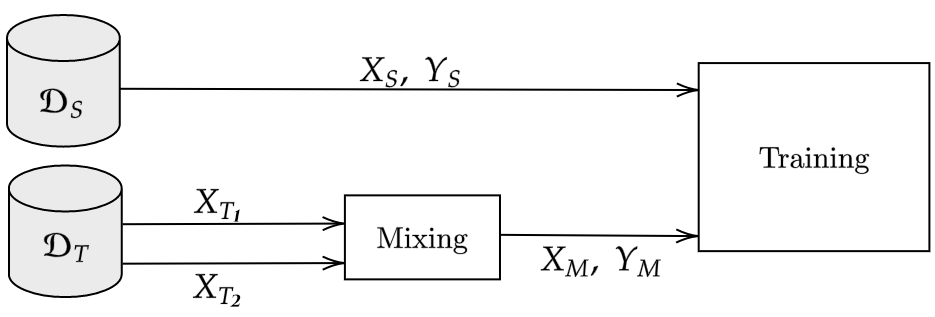}
    \caption{Diagram showing Naive Mixing of images. Images $X_{T_1}$ and $X_{T_2}$ from the target dataset $\mathcal D_T$ are mixed, generating the augmented image $X_M$ and its pseudo-label $Y_M$. The segmentation network is trained on batches of augmented images and batches of images from the source dataset $X_S$ (for which ground-truth labels, $Y_S$, are available).}
    \label{fig:naive_classmix}
\end{figure}

As stated previously, in the original formulation of consistency regularization based on mixing for SSL \cite{French,StructuredLoss,ClassMix}, samples are mixed within the unlabelled dataset to generate augmented images. In UDA, the unlabelled samples are the ones from the target dataset, so a natural adaptation to this context would be to mix target-domain images. This approach, henceforth referred to as ``Naive Mixing'', mixes target-domain samples to generate augmented images and corresponding pseudo-labels and then trains the network using both the augmented images and the source-domain images, as illustrated in Figure \ref{fig:naive_classmix}. 

However, this intuitive adaptation performs poorly in practice, as shown in the experiments in Section \ref{sec:results}. The resulting segmentation network conflates some of the classes when predicting the semantics of target-domain images. For instance, classes with fewer occurrences like 'sidewalk' are confused with more frequent and semantically similar classes, like 'road'. Similarly, the 'rider' class is misclassified as 'person', and 'terrain' as 'vegetation', among other classes. This seems to be a consistent pattern across different seeds of training, and impacts performance considerably. The problem occurs exclusively for the target domain images, not for the images in the source domain.


This problem, which we refer to as class conflation, is similar to one identified in early works applying pseudo-labelling to UDA for semantic segmentation tasks \cite{zou2018domain}, where they point out a bias towards easy-to-transfer classes when applying pseudo-labelling naively to UDA. 
Using pseudo-labelling in a consistency regularization setting combined with mixing, as used for SSL in \cite{ClassMix}, inherits the same underlying issues.
While existing works have proposed other improvements for how to correct erroneous pseudo-label generation arising due to the domain shift \cite{zou2018domain,zou2019confidence,zheng2020rectifying}, we instead propose a change in the augmentation procedure, detailed in the next subsection.

\subsection{DACS}
\label{sec:DACS}
\begin{figure}[t]
    \centering
    \includegraphics[width=\columnwidth]{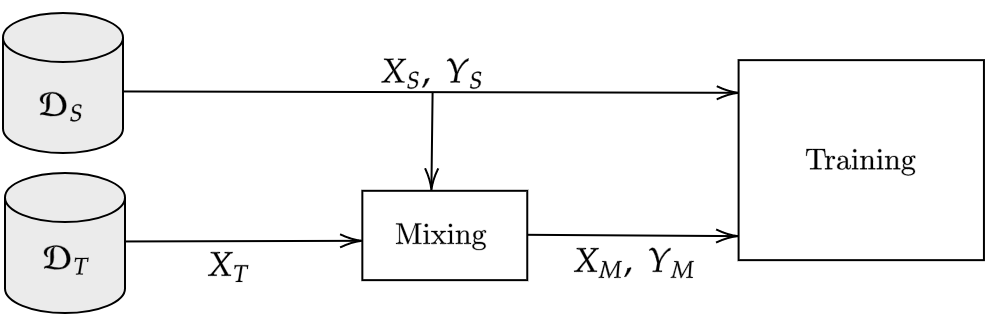}
    \caption{Diagram showing DACS. The images $X_S$ and $X_T$ are mixed together, using $Y_S$ for the labels of $X_S$, instead of a predicted semantic map to determine the binary mask. The segmentation network is then trained on both batches of augmented images and images from the source dataset.}
    \label{fig:dacs}
\end{figure}

\begin{figure}[t]
    \centering
    \includegraphics[width=\columnwidth]{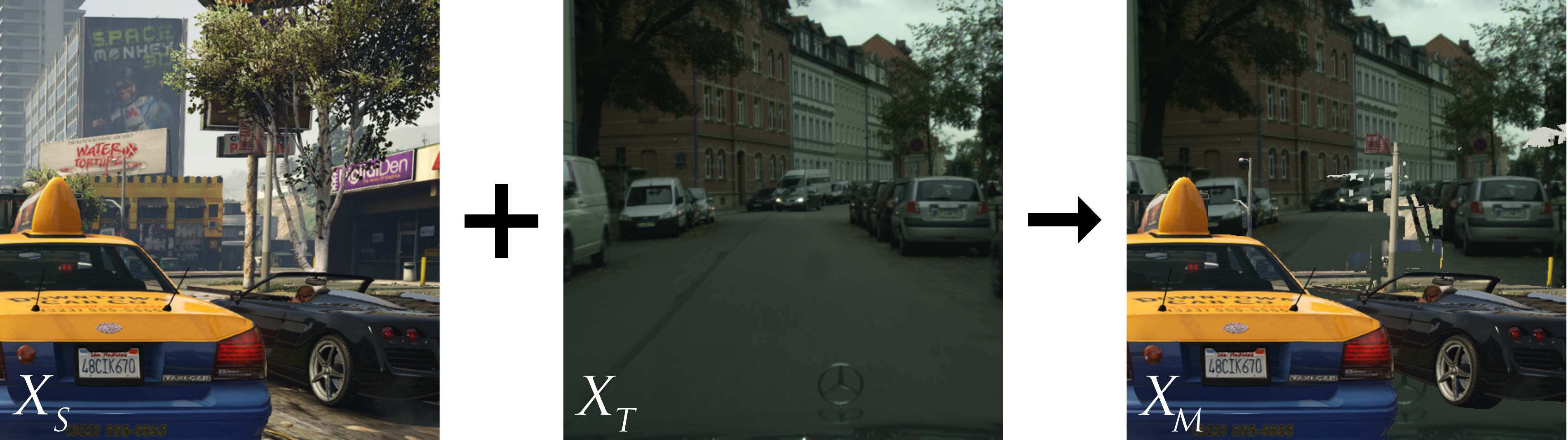}
    \caption{Example augmentation used in DACS: an image from the source domain (in this case synthetic data from GTA5) is mixed with an image from the target domain (Cityscapes), resulting in an augmented image which contains parts from both domains.}
    \label{fig:crossmix_example_image}
\end{figure}

Our proposed solution builds on the idea of mixing images across domains. New, augmented samples are created by having one set of pixels coming from a source domain image, and one set of pixels coming from a target domain image. These new samples are then used to train the network. By default we are using the mixing strategy ClassMix \cite{ClassMix}, but any mixing method based on binary masks would be valid. In our adaptation of ClassMix, half of the classes in a source domain image are selected, and the corresponding pixels are cut out and pasted onto an image from the target domain. To construct the corresponding image with pseudo-labels, the target domain image is also run through the network before the mixing in order to produce a pseudo-label for it. This pseudo-label is then mixed in the same way as the image, together with the corresponding ground-truth label from the source domain image. A diagram explaining the mixing procedure is shown in Figure \ref{fig:dacs}.
An example of the input images and the resulting augmentation is illustrated in Figure \ref{fig:crossmix_example_image}. Note that the resulting mixed images are not necessarily realistic. However, complete realism of the augmentations is not required for the functioning of our method, as shown by the results in Section \ref{sec:results}.

The proposed cross-domain mixing greatly improves performance compared to Naive Mixing.
Before introducing cross-domain mixing, we have two types of data: 1) source domain data with ground-truth labels, and 2) target domain data with potentially conflated pseudo-labels, where the gap between the domains may be large. Due to this potentially large gap, a network may implicitly learn to discern between the domains in order to perform better in the task, and (incorrectly) learn that the class distributions are very different in the two domains. With cross-domain mixing, we introduce new data, and considering that the labels for these new images partly come from the source domain, they will not be conflated for entire images. Furthermore, the pixels that are pseudo-labelled (target-domain labels) and the pixels that have ground-truth labels (source-domain labels) may now be neighbors in an image, making the implicit discerning between domains unlikely, since it would have to be done at a pixel level. Both of these aspects help the network to better deal with the domain gap, and effectively solve the class conflation problem, as shown in Section \ref{sec:results}, resulting in considerably better performance. The overall UDA algorithm that trains on source-domain images and cross-domain augmentations is what we refer to as DACS. 

The implementation of DACS is presented as pseudocode in Algorithm \ref{alg:crossmix}, where the source-domain and target-domain datasets are referred to as $D_S$ and $D_T$, respectively. A batch of images and labels,  $X_S$ and $Y_S$, is sampled from $\mathcal D_S$, and a batch of images, $X_T$, from $\mathcal D_T$. The images in $X_T$ are then fed to the network $f_\theta$, which outputs their predicted semantic maps $\hat{Y}_T$. Then, the augmented images $X_M$ are created by mixing $X_S$ and $X_T$, and the pseudo-labels $Y_M$ by mixing the corresponding maps in $Y_S$ and $\hat{Y}_T$. From this point forward, the algorithm resembles a supervised learning approach: compute predictions, compare them with the labels (in our case using the cross-entropy loss), perform backprogragation, and perform a step of gradient descent. This process is then repeated for a predetermined amount of iterations $N$.

\begin{algorithm}[t]
\caption{DACS algorithm}
\label{alg:crossmix}
\begin{algorithmic}[1]
\Require Source-domain and target-domain datasets $\mathcal D_S$ and $\mathcal D_T$, segmentation network $f_\theta$. 

\State Initialize network parameters $\theta$ randomly.
\For{$i=1$ to $N$}

    \State $X_S, Y_S\sim \mathcal D_S$ 
    \State $X_T \sim \mathcal D_T$ 
    \State $\hat{Y}_T \gets f_{\theta}(X_T)$ 
    \State $X_M, Y_M\gets$ Augmentation and pseudo-label from mixing $X_S, Y_S$, $X_T$ and $\hat{Y}_T$.
    
    \State $\hat{Y}_S\gets f_\theta(X_S),~ \hat{Y}_M\gets f_\theta(X_M)$ \Comment{Compute predictions.}
    
    \State $\ell\gets 
    L(\hat{Y}_S, Y_S, \hat{Y}_M, Y_M)$ \Comment{Compute loss for the batch.}
    \State Compute $\nabla_\theta \ell$ by backpropagation (treating $Y_M$ as constant.)
    
    \State Perform one step of stochastic gradient descent on $\theta$.

\EndFor
\State \Return $f_\theta$
\end{algorithmic}
\end{algorithm}

\subsection{Loss Function}
\label{sec:loss_and_training}
In DACS, the network parameters $\theta$ are trained by minimizing the following loss:
\begin{equation}
    \mathcal L(\theta) = \mathbb E\left[H\Big(f_\theta(X_S), Y_S\Big) + \lambda H\Big(f_\theta(X_M), Y_M\Big)\right].
\end{equation}
where the expectation is over batches of random variables; $X_S$, $Y_S$, $X_M$, and $Y_M$. Images in $X_S$ are sampled uniformly from the source domain distribution, and $Y_S$ is the corresponding labels. Further, $X_M$ and $Y_M$ are the mixed images and pseudo-labels, created by performing cross-domain mixing between the images sampled uniformly at random from the source domain and those from the target domain, as explained above. Lastly, $H$ is the cross-entropy between the predicted semantic maps and the corresponding labels (ground-truth or pseudo) averaged over all images and pixels, and $\lambda$ is a hyper-parameter that decides how much the unsupervised part of the loss affects the overall training. In line with \cite{French,ClassMix}, we use an adaptive schedule for $\lambda$, where it is the proportion of pixels, for each image, where the predictions of $f_\theta$ on that image have a confidence above a certain threshold. Training is performed by stochastic gradient descent on the loss using batches with the same number of source-domain images and augmented images.

\section{Experiments}
\label{sec:results}
In order to validate the proposed DACS algorithm, we evaluate it in two popular datasets for UDA and compare to the state of the art for these tasks. We also describe additional experiments related to the source of the class conflation problem, as well as the effect of the choice of mixing strategy. This section details the experimental setup and provides the qualitative and quantitative results found.

\vspace{0.4cm}
\noindent{\bf Implementation Details.}
For all the experiments in this paper, we adopt the widely used \cite{tsai2018learning,luo2019significanceaware,luo2018taking,yang2019adversarial,tsai2019domain,vu2018advent,zhang2018fully,zou2018domain,zou2019confidence,li2019bidirectional,yang2020contextaware,zheng2019unsupervised,zheng2020rectifying, yang2020fda, wang2020differential, lv2020PIT, DBLP:journals/corr/abs-2008-12197, yang2020labelddriven, DBLP:conf/cvpr/KimB20a, DBLP:conf/nips/ZhangZ0T19} DeepLab-v2 framework \cite{DeepLabv2} with a ResNet101 backbone \cite{ResNet} as our model. The backbone is pretrained on ImageNet \cite{imagenet} and on MSCOCO \cite{mscoco}. Most hyper-parameters are identical to those used in \cite{tsai2018learning}. We use Stochastic Gradient Descent with Nesterov acceleration, and an initial learning rate of 2.5$\times10^{-4}$, which is then decreased using polynomial decay with exponent 0.9 as in \cite{DeepLabv2}. Weight decay is set to 5$\times10^{-4}$ and momentum to 0.9. Source images are rescaled to $760\times1280$ and target images to $512\times1024$, after which random crops of size $512\times512$ are extracted. For our main results we use ClassMix unless otherwise stated and in addition we also apply Color jittering and Gaussian blurring on the mixed images. We train using batches with 2 source images and 2 mixed images for 250k iterations. The code is implemented using PyTorch, and is available at \url{https://github.com/vikolss/DACS}. Experiments were performed using a GTX 1080 Ti GPU with 12 GB memory.

\begin{figure}[t]
    \centering
    \begin{adjustbox}{width=\columnwidth}
    \begin{subfigure}[b]{1.2\textwidth}
        \caption*{\centering \fontsize{50}{30}\selectfont Cityscapes}
        \includegraphics[width=\textwidth]{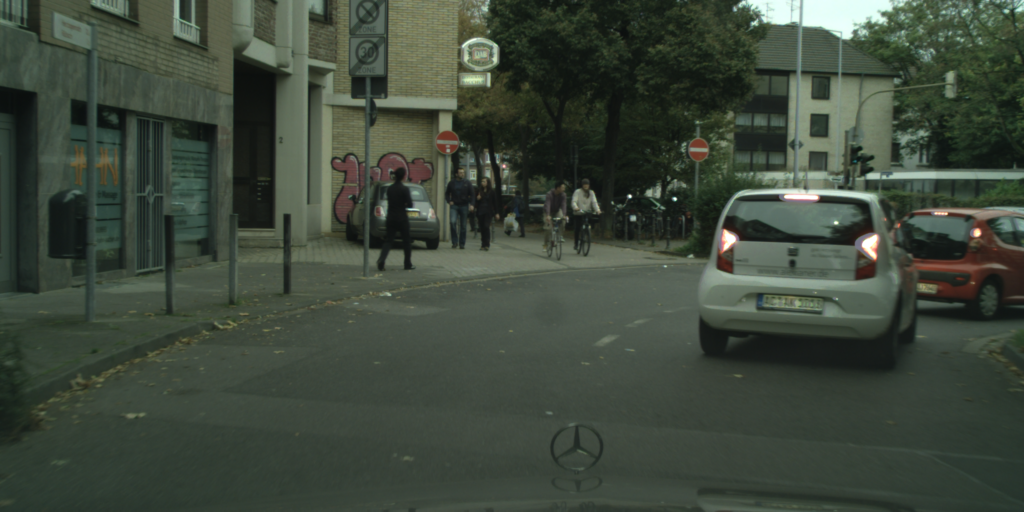}
    \end{subfigure}
    \hspace{0.01cm}
    \begin{subfigure}[b]{1.2\textwidth}
        \caption*{\centering \fontsize{50}{30}\selectfont GTA5}
        \includegraphics[width=\textwidth]{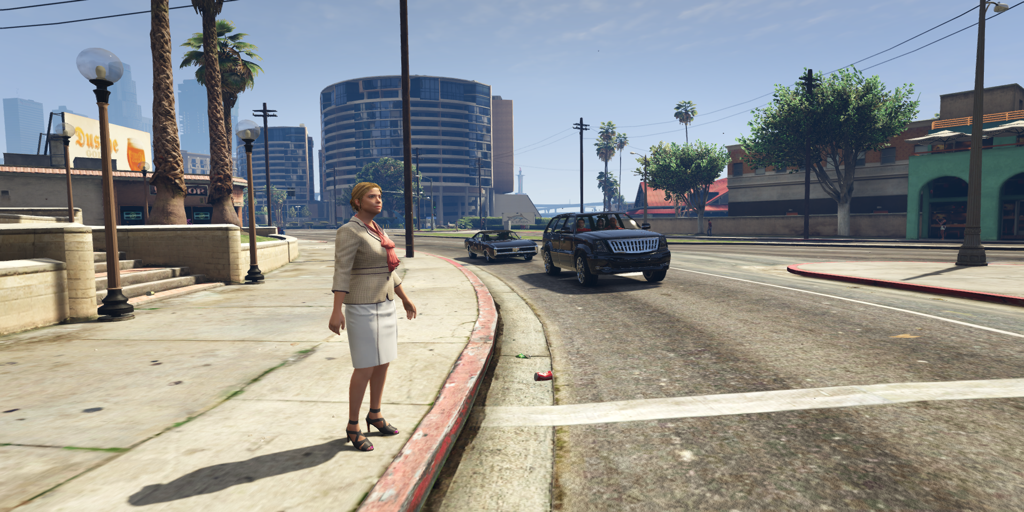}
    \end{subfigure}
    \hspace{0.01cm}
    \begin{subfigure}[b]{1.2\textwidth}
        \caption*{\centering \fontsize{50}{30}\selectfont SYNTHIA}
        \includegraphics[width=\textwidth]{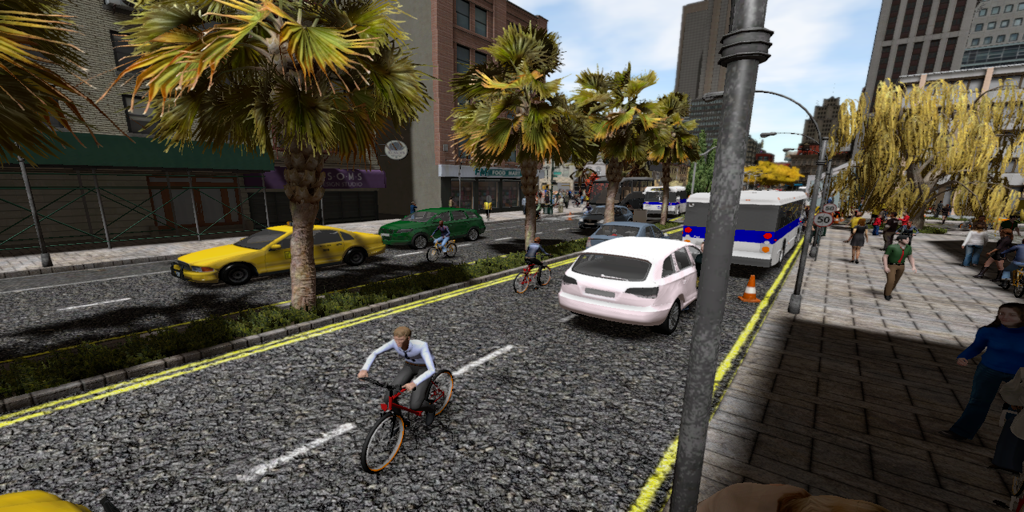}
    \end{subfigure}
    \end{adjustbox}
    \begin{adjustbox}{width=\columnwidth}
    \begin{subfigure}[b]{1.5\textwidth}
        \includegraphics[width=\textwidth]{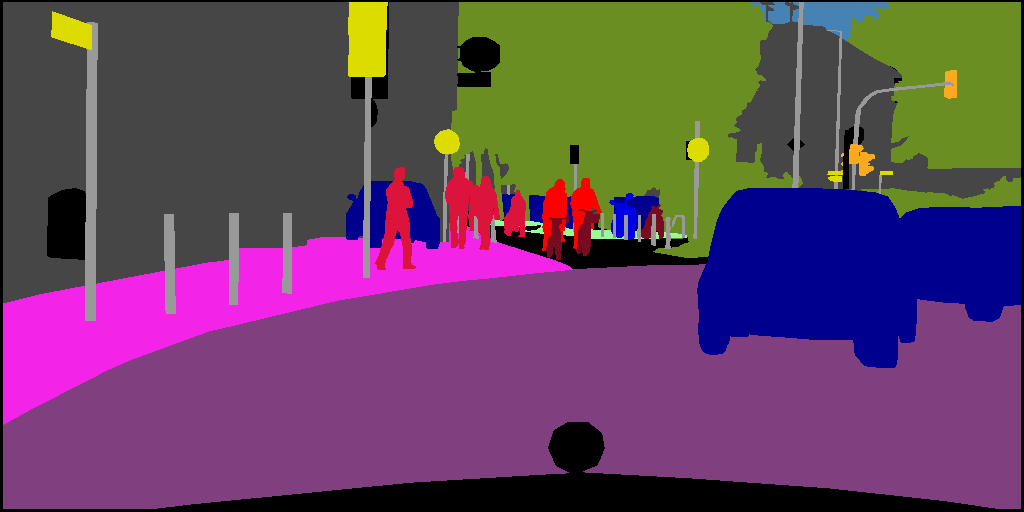}
    \end{subfigure}
    \hspace{0.01cm}
    \begin{subfigure}[b]{1.5\textwidth}
        \includegraphics[width=\textwidth]{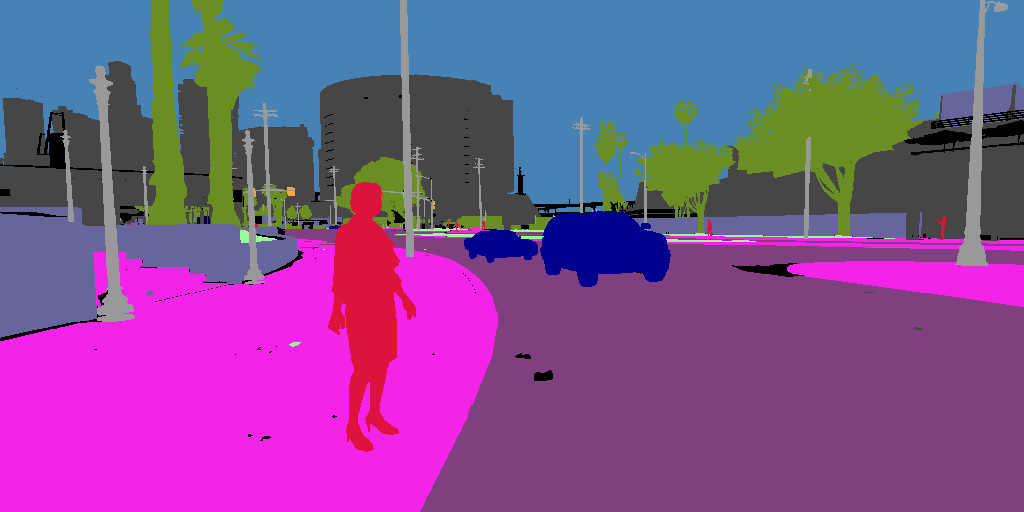}
    \end{subfigure}
    \hspace{0.01cm}
    \begin{subfigure}[b]{1.5\textwidth}
        \includegraphics[width=\textwidth]{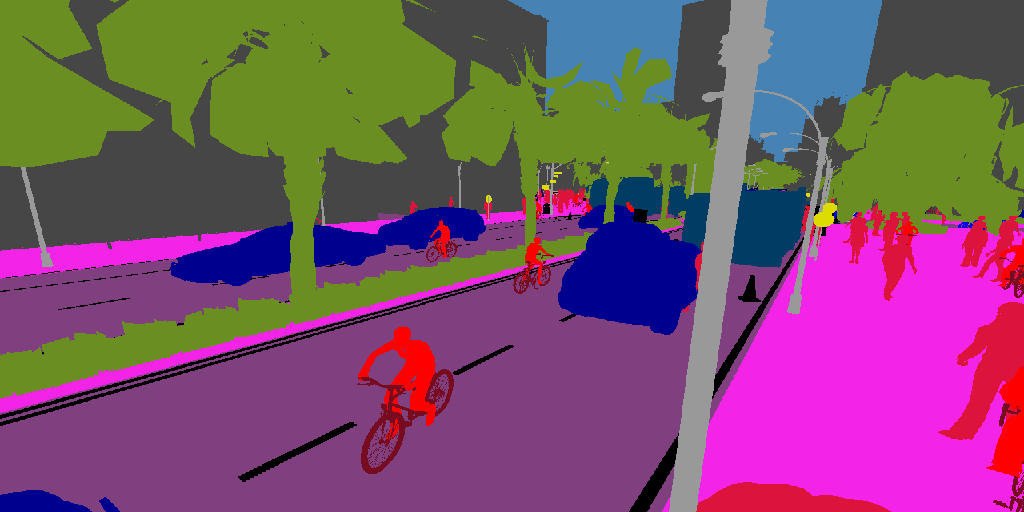}
    \end{subfigure}
    \end{adjustbox}
    
    \caption{Images from the Cityscapes, GTA5, and SYNTHIA datasets along with their corresponding semantic maps.}
    \label{fig:exampleimages}
\end{figure}

\vspace{0.4cm}
\noindent{\bf Datasets.} We present results for two synthetic-to-real benchmarks common for UDA for semantic segmentation. Namely GTA5 $\rightarrow$ Cityscapes and SYNTHIA $\rightarrow$ Cityscapes. The target dataset Cityscapes has 2,975 training images taken from a car in urban environments and is labelled with 19 classes \cite{Cityscapes}. The source datasets GTA5 \cite{GTAdataset} and SYNTHIA \cite{SYNTHIAdataset} contain 24,966 and 9,400 synthetic training images respectively. Example images of all three datasets are shown in Figure \ref{fig:exampleimages} together with ground-truth semantic maps. The GTA5 images are labelled with the same 19 classes as Cityscapes whereas the SYNTHIA data is labelled with 16 of the 19 classes. 
All results are reported with the Intersection over Union (IoU) metric per class and the mean Intersection over Union (mIoU) over all classes, the standard performance metric for semantic segmentation.

\begin{table*}[t]
    \centering
    \caption{GTA5 to Cityscapes, our results are averages from three runs, and shown as per-class IoU and mIoU. We also compare to several previous works. We present results for DACS, as well as for training only on data from the source domain.}
    \begin{adjustbox}{width=1\textwidth}
    \begin{tabular}{l|lllllllllllllllllll|l}
    \hline
                Method & Road & SW & Build & Wall & Fence & Pole & TL & TS & Veg & Terrain & Sky  & Person & Rider & Car  & Truck & Bus  & Train & MC & Bike & mIoU \\ \hline
                
AdaptSegNet \cite{tsai2018learning}   & 86.5 & 36.0     & 79.9     & 23.4 & 23.3  & 23.9 & 35.2          & 14.8         & 83.4         & 33.3    & 75.6 & 58.5   & 27.6  & 73.7 & 32.5  & 35.4 & 3.9   & 30.1       & 28.1    & 42.4       \\ 
SIBAN \cite{luo2019significanceaware}   & 88.5 & 35.4 & 79.5 & 26.3 & 24.3 & 28.5 & 32.5 & 18.3 & 81.2 & 40.0 & 76.5 & 58.1 & 25.8 & 82.6 & 30.3 & 34.4 & 3.4 & 21.6 & 21.5  & 42.6  \\ 
CLAN \cite{luo2018taking}   & 87.0 & 27.1 & 79.6 & 27.3 & 23.3 & 28.3 & 35.5 & 24.2 & 83.6 & 27.4 & 74.2 & 58.6 & 28.0 & 76.2 & 33.1 & 36.7 & 6.7 & 31.9 & 31.4 & 43.2 \\ 
APODA \cite{yang2019adversarial}   & 85.6 & 32.8 & 79.0 & 29.5 & 25.5 & 26.8 & 34.6 & 19.9 & 83.7 & 40.6 & 77.9 & 59.2 & 28.3 & 84.6 & 34.6 & 49.2 & 8.0 & 32.6 & 39.6 & 45.9 \\ 
PatchAlign \cite{tsai2019domain}   & 92.3 & 51.9 & 82.1 & 29.2 & 25.1 & 24.5 & 33.8 & 33.0 & 82.4 & 32.8 & 82.2 & 58.6 & 27.2 & 84.3 & 33.4 & 46.3 & 2.2 & 29.5 & 32.3 & 46.5 \\ 
AdvEnt \cite{vu2018advent}   & 89.4 & 33.1 & 81.0 & 26.6 & 26.8 & 27.2 & 33.5 & 24.7 & 83.9 & 36.7 & 78.8 & 58.7 & 30.5 & 84.8 & 38.5 & 44.5 & 1.7 & 31.6 & 32.4 & 45.5 \\ 
FCAN \cite{zhang2018fully}   & - & - & - & - & - & - & - & - & - & - & - & - & - & - & - & - & - & - & - & 46.6 \\ 
CBST \cite{zou2018domain}   & 91.8 & 53.5 & 80.5 & 32.7 & 21.0 & 34.0 & 28.9 & 20.4 & 83.9 & 34.2 & 80.9 & 53.1 & 24.0 & 82.7 & 30.3 & 35.9 & 16.0 & 25.9 & 42.8 & 45.9 \\ 

MRKLD-SP \cite{zou2019confidence} & 90.8 & 46.0 & 79.9 & 27.4 & 23.3 & \textbf{42.3} & 46.2 & 40.9 & 83.5 & 19.2 & 59.1 & 63.5 & 30.8 & 83.5 & 36.8 & 52.0 & 28.0 & \textbf{36.8} & 46.4 & 49.2
  \\ 

BDL \cite{li2019bidirectional}   & 91.0 & 44.7 & 84.2 & 34.6 & 27.6 & 30.2 & 36.0  & 36.0 & 85.0 & 43.6 & 83.0 & 58.6 & 31.6 & 83.3 & 35.3 & 49.7 & 3.3 & 28.8 & 35.6 & 48.5 \\ 

CADASS \cite{yang2020contextaware}   & 91.3 & 46.0 & 84.5 & 34.4 & 29.7 & 32.6 & 35.8 & 36.4 & 84.5 & 43.2 & 83.0 & 60.0 & 32.2 & 83.2 & 35.0 & 46.7 & 0.0 & 33.7 & 42.2 & 49.2 \\ 

MRNet \cite{zheng2019unsupervised}   & 89.1 & 23.9 & 82.2 & 19.5 & 20.1 & 33.5 & 42.2 & 39.1 & 85.3 & 33.7 & 76.4 & 60.2 & 33.7 & 86.0 & 36.1 & 43.3 & 5.9 & 22.8 & 30.8 & 45.5 \\ 
R-MRNet \cite{zheng2020rectifying}   & 90.4 & 31.2 & 85.1 & 36.9 & 25.6 & 37.5 & \textbf{48.8} & 48.5 & 85.3 & 34.8 & 81.1 & 64.4 & 36.8 & 86.3 & 34.9 & 52.2 & 1.7 & 29.0 & 44.6 & 50.3 \\ 

PIT \cite{lv2020PIT} & 87.5 & 43.4 & 78.8 & 31.2 & 30.2 & 36.3 & 39.9 & 42.0 & 79.2 & 37.1 & 79.3 & 65.4 & \textbf{37.5} & 83.2 & \textbf{46.0} & 45.6 & 25.7 & 23.5 & \textbf{49.9} & 50.6 \\ 

SIM \cite{wang2020differential} & 90.6 & 44.7 & 84.8 & 34.3 & 28.7 & 31.6 & 35.0 & 37.6 & 84.7 & 43.3 & 85.3 & 57.0 & 31.5 & 83.8 & 42.6 & 48.5 & 1.9 & 30.4 & 39.0 & 49.2 \\ 

FDA \cite{yang2020fda} & 92.5 & 53.3 & 82.4 & 26.5 & 27.6 & 36.4 & 40.6 & 38.9 & 82.3 & 39.8 & 78.0 & 62.6 & 34.4 & 84.9 & 34.1 & \textbf{53.1} & 16.9 & 27.7 & 46.4 & 50.45 \\ 

Yang et al. \cite{yang2020labelddriven} & 90.8 & 41.4 & 84.7 & 35.1 & 27.5 & 31.2 & 38.0 & 32.8 & 85.6 & 42.1 & 84.9 & 59.6 & 34.4 & 85.0 & 42.8 & 52.7 & 3.4 & 30.9 & 38.1 & 49.5 \\

Kim et al. \cite{DBLP:conf/cvpr/KimB20a} & 92.9 & 55.0 & 85.3 & 34.2 & 31.1 & 34.9 & 40.7 & 34.0 & 85.2 & 40.1 & 87.1 & 61.0 & 31.1 & 82.5 & 32.3 & 42.9 & 0.3 & 36.4 & 46.1 & 50.2 \\ 

CAG-UDA \cite{DBLP:conf/nips/ZhangZ0T19} & 90.4 & 51.6 & 83.8 & 34.2 & 27.8 & 38.4 & 25.3 & 48.4 & 85.4 & 38.2 & 78.1 & 58.6 & 34.6 & 84.7 & 21.9 & 42.7 & \textbf{41.1} & 29.3 & 37.2 & 50.2 \\ 

IAST \cite{DBLP:journals/corr/abs-2008-12197} & \textbf{93.8} & \textbf{57.8} & 85.1 & \textbf{39.5} & 26.7 & 26.2 & 43.1 & 34.7 & 84.9 & 32.9 & 88.0 & 62.6 & 29.0 & \textbf{87.3} & 39.2 & 49.6 & 23.2 & 34.7 & 39.6 & 51.5 \\

\hline

Source   & 63.31 & 15.65 & 59.39 & 8.56 & 15.17 & 18.31 & 26.94 & 15.00 & 80.46 & 15.25 & 72.97 & 51.04 & 17.67 & 59.68 & 28.19 & 33.07 & 3.53 & 23.21 & 16.73 & 32.85 \\


DACS& 89.90 & 39.66 & \textbf{87.87} & 30.71 & \textbf{39.52} & 38.52 & 46.43 & \textbf{52.79} & \textbf{87.98} & \textbf{43.96} & \textbf{88.76} & \textbf{67.20} & 35.78 & 84.45 & 45.73 & 50.19 & 0.00 & 27.25 & 33.96 & \textbf{52.14} \\ \hline

    \end{tabular}
    \end{adjustbox}
    \label{tab:GTACSresults}
\end{table*}{}

\subsection{GTA5 $\rightarrow$ Cityscapes Results} We present our results for GTA5 $\rightarrow$ Cityscapes in Table \ref{tab:GTACSresults}, alongside the results from several existing works on the same task. All of our comparisons have results for the same DeepLab-v2 network, but for completeness we choose to include their best presented performance, regardless of backbone. In the table, ``Source'' refers to a model only trained on the source data and then evaluated on the target data, and serves as our baseline, and DACS refers to the results from our proposed method, which achieves the strongest results for seven of the individual classes, as well as an overall mIoU of 52.14\%, higher than all previous methods, effectively pushing the state of the art for this task.

\begin{figure*}[th!]
    \centering
    \begin{adjustbox}{width=1\textwidth}
    \begin{subfigure}[b]{\textwidth}
        \caption*{\centering \fontsize{40}{60}\selectfont Image}
        \includegraphics[width=\textwidth]{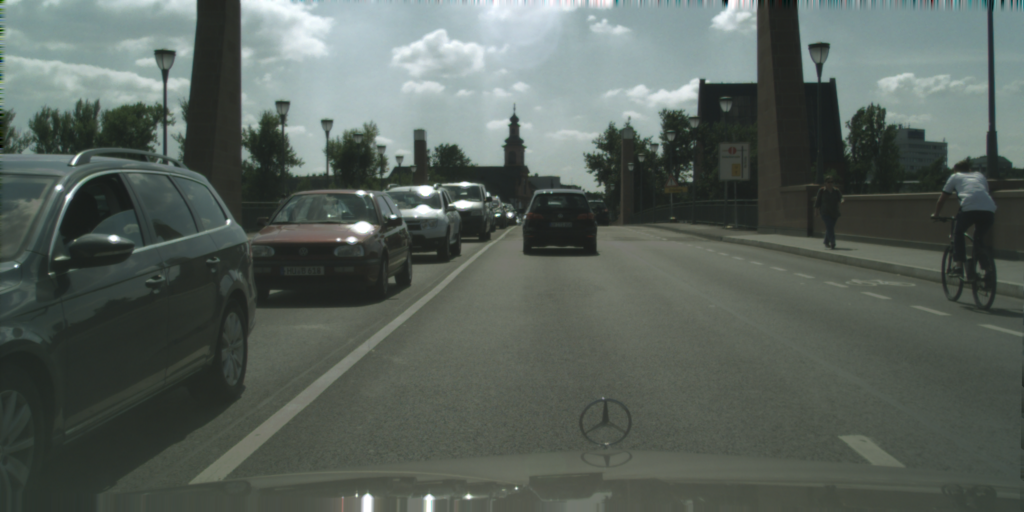}
    \end{subfigure}
    \hspace{0.01cm}
    \begin{subfigure}[b]{\textwidth}
        \caption*{\centering \fontsize{40}{60}\selectfont Ground-truth}
        \includegraphics[width=\textwidth]{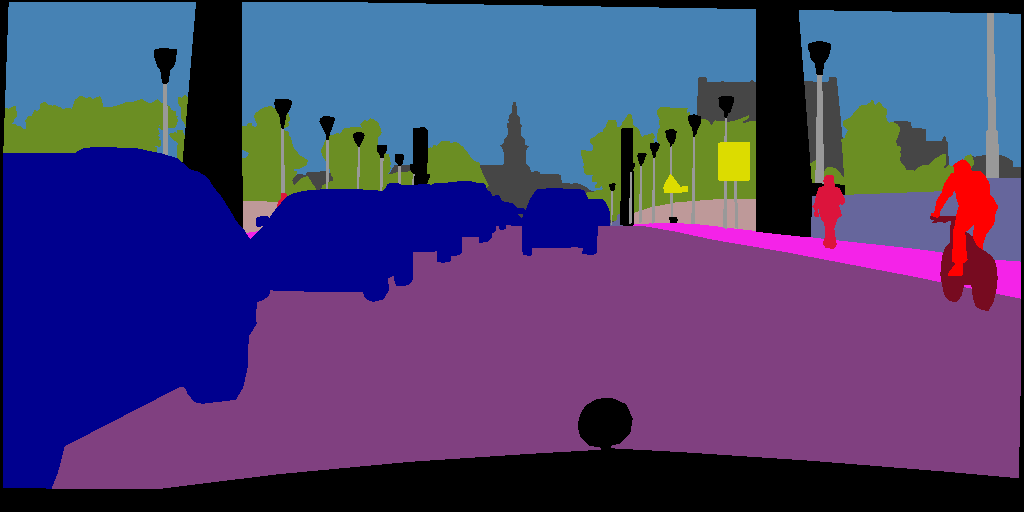}
    \end{subfigure}
    \hspace{0.01cm}
    \begin{subfigure}[b]{\textwidth}
        \caption*{\centering \fontsize{40}{60}\selectfont Source}
        \includegraphics[width=\textwidth]{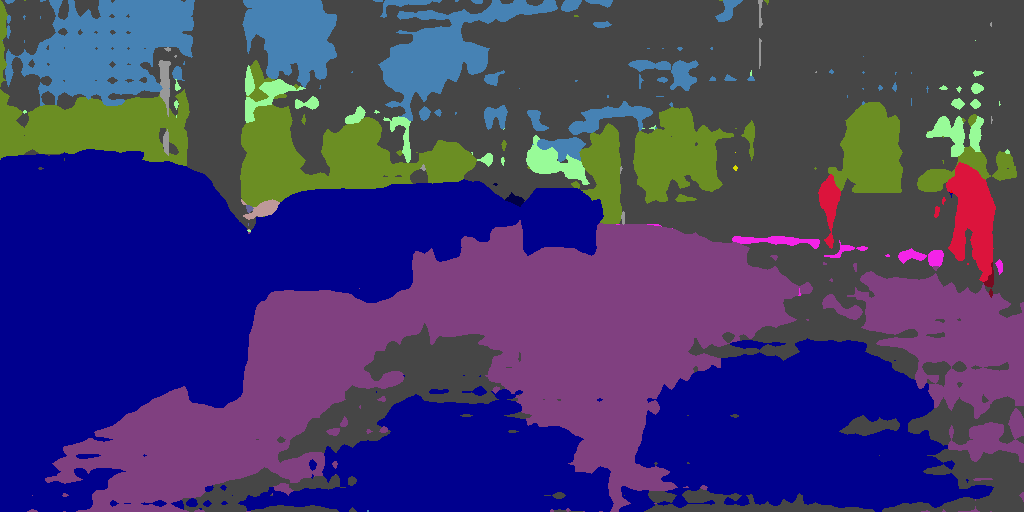}
    \end{subfigure}
    \hspace{0.01cm}
    \begin{subfigure}[b]{\textwidth}
        \caption*{\centering \fontsize{40}{60}\selectfont Naive Mixing}
        \includegraphics[width=\textwidth]{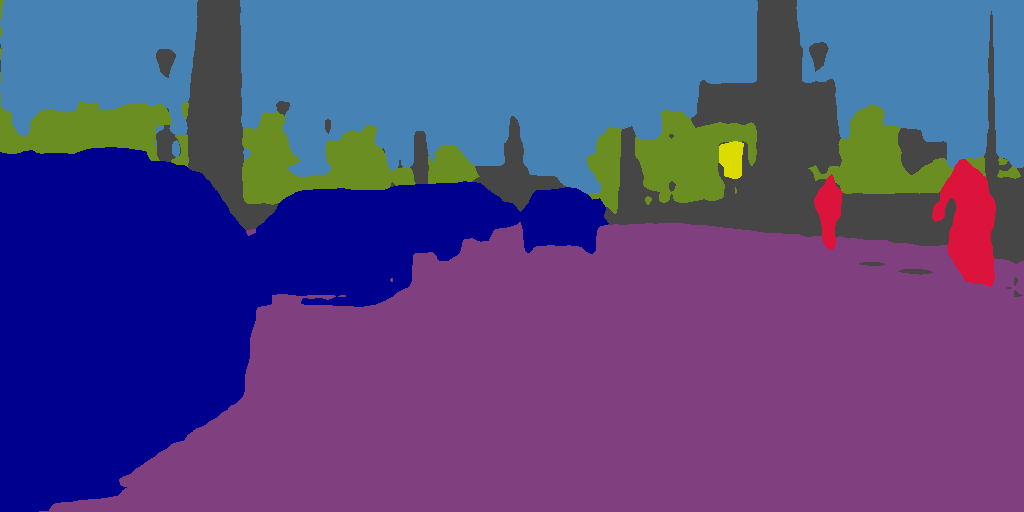}
    \end{subfigure}
    \hspace{0.01cm}
    \begin{subfigure}[b]{\textwidth}
        \caption*{\centering \fontsize{40}{60}\selectfont DACS}
        \includegraphics[width=\textwidth]{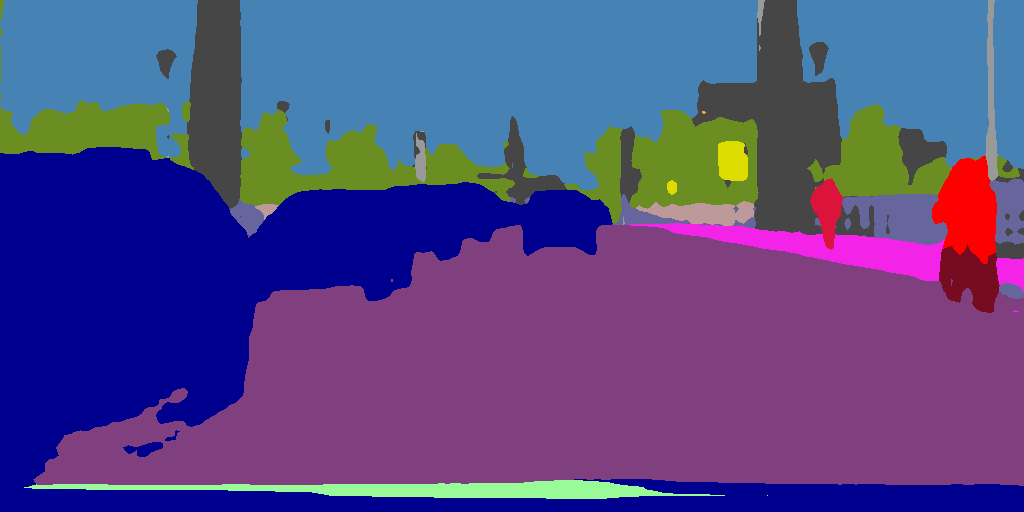}
    \end{subfigure}
    \end{adjustbox}
    
    \begin{adjustbox}{width=1\textwidth}
    \begin{subfigure}[b]{\textwidth}
        \includegraphics[width=\textwidth]{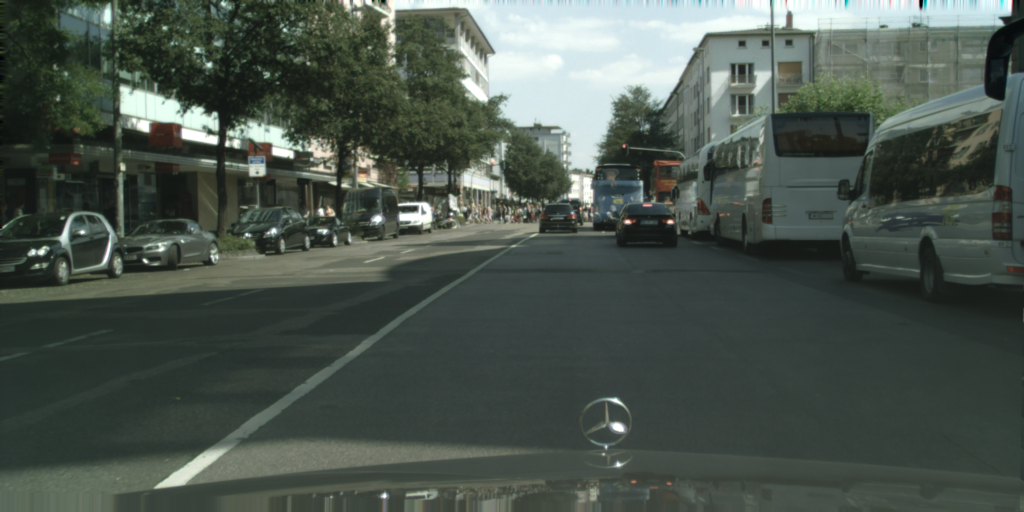}
    \end{subfigure}
    \hspace{0.01cm}
    \begin{subfigure}[b]{\textwidth}
        \includegraphics[width=\textwidth]{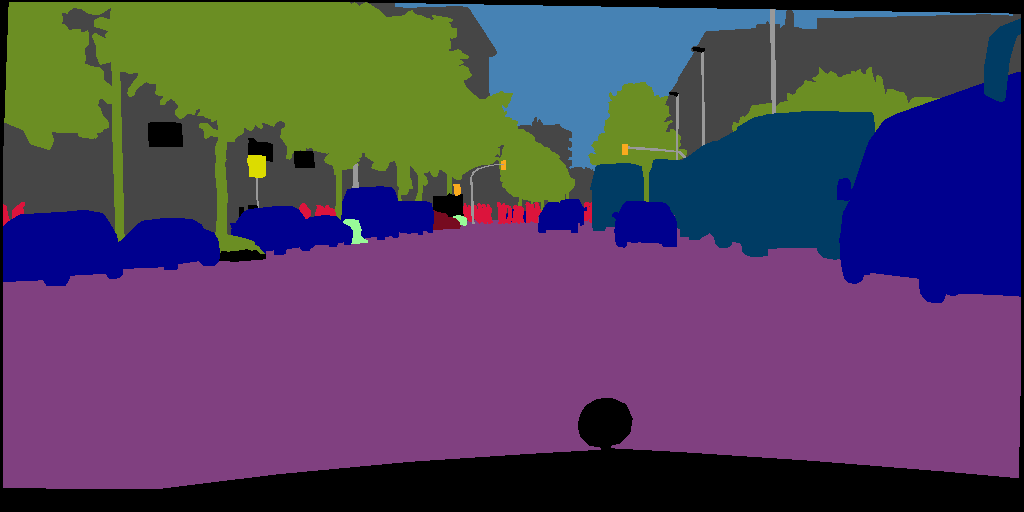}
    \end{subfigure}
    \hspace{0.01cm}
    \begin{subfigure}[b]{\textwidth}
        \includegraphics[width=\textwidth]{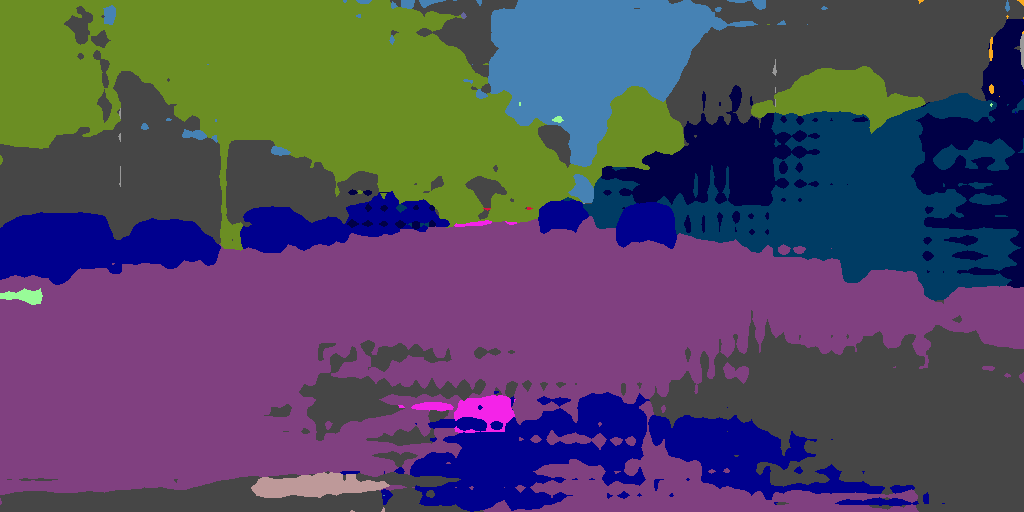}
    \end{subfigure}
    \hspace{0.01cm}
    \begin{subfigure}[b]{\textwidth}
        \includegraphics[width=\textwidth]{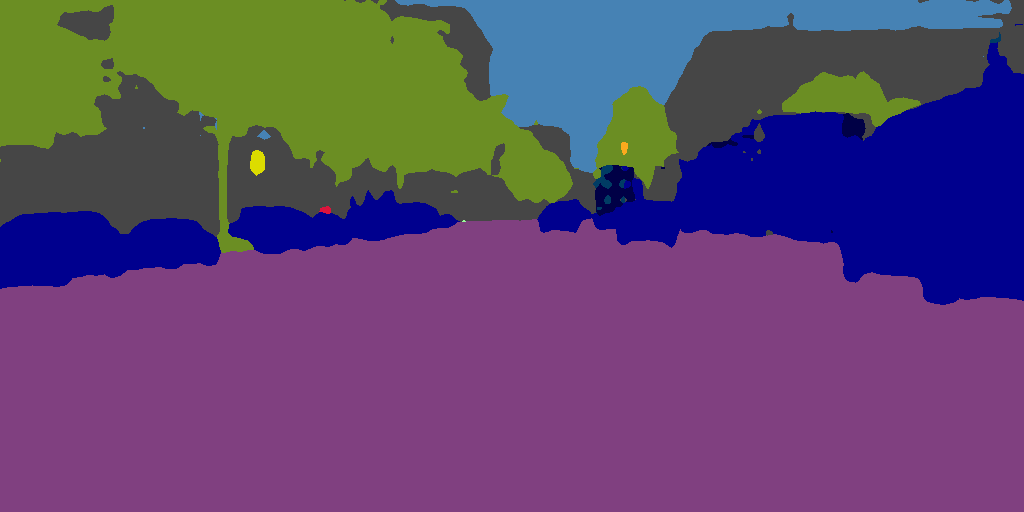}
    \end{subfigure}
    \hspace{0.01cm}
    \begin{subfigure}[b]{\textwidth}
        \includegraphics[width=\textwidth]{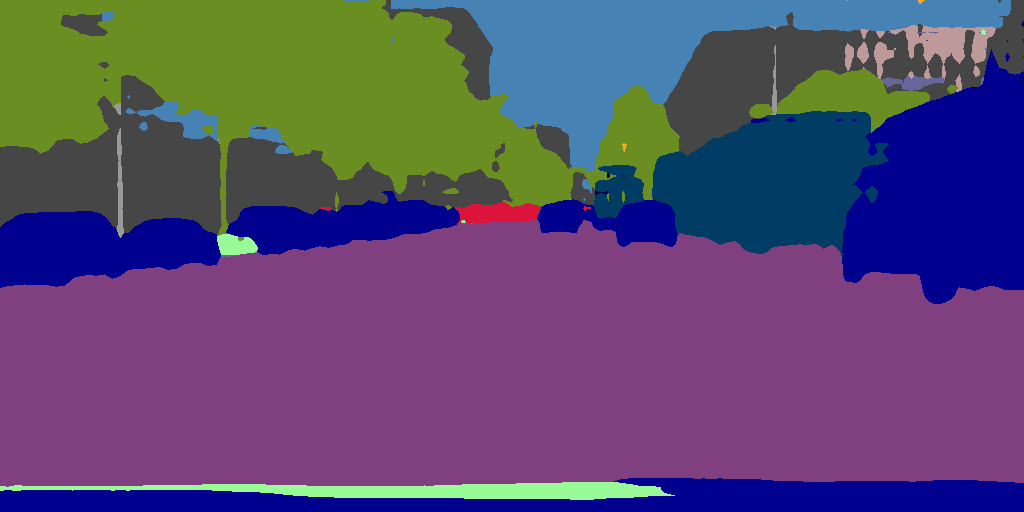}
    \end{subfigure}
    \end{adjustbox}
    
    
    \begin{adjustbox}{width=1\textwidth}
    \begin{subfigure}[b]{\textwidth}
        \includegraphics[width=\textwidth]{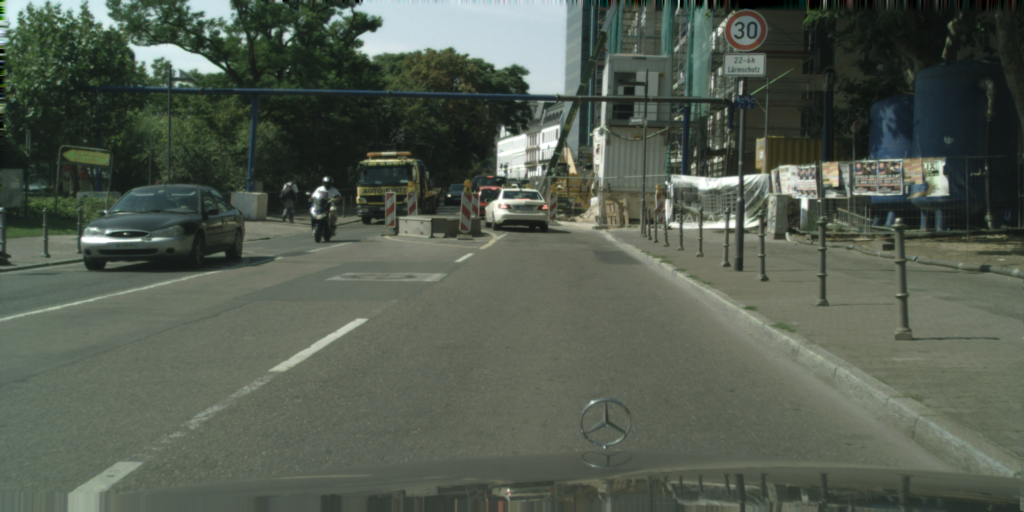}
    \end{subfigure}
    \hspace{0.01cm}
    \begin{subfigure}[b]{\textwidth}
        \includegraphics[width=\textwidth]{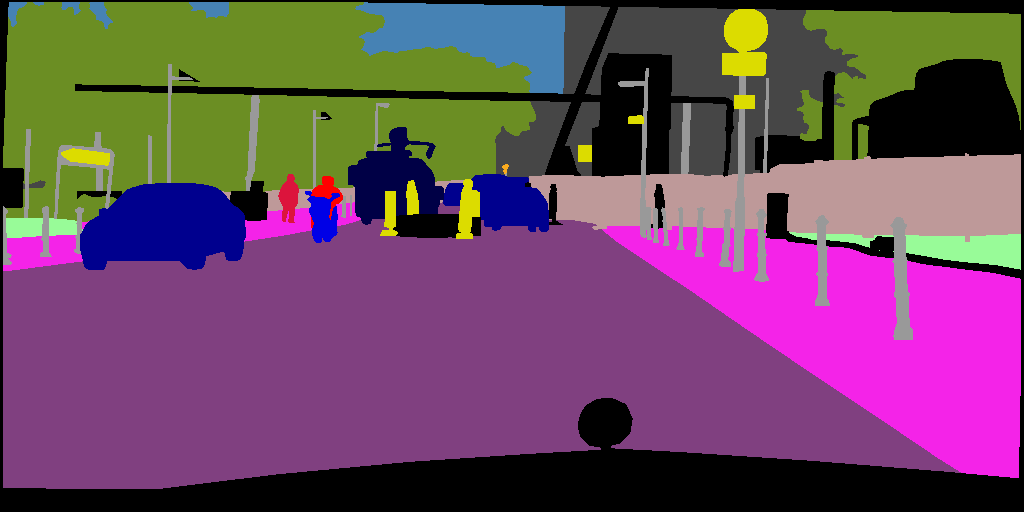}
    \end{subfigure}
    \hspace{0.01cm}
    \begin{subfigure}[b]{\textwidth}
        \includegraphics[width=\textwidth]{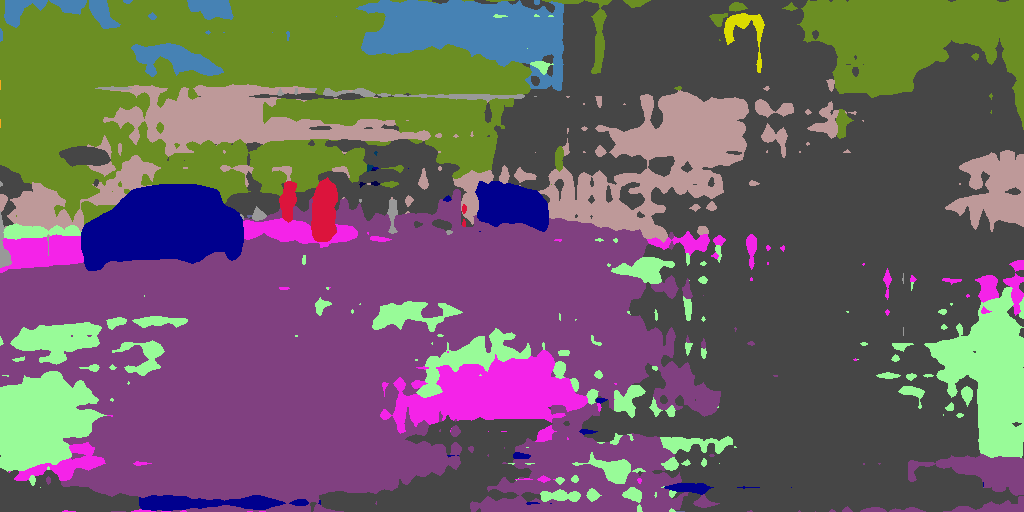}
    \end{subfigure}
    \hspace{0.01cm}
    \begin{subfigure}[b]{\textwidth}
        \includegraphics[width=\textwidth]{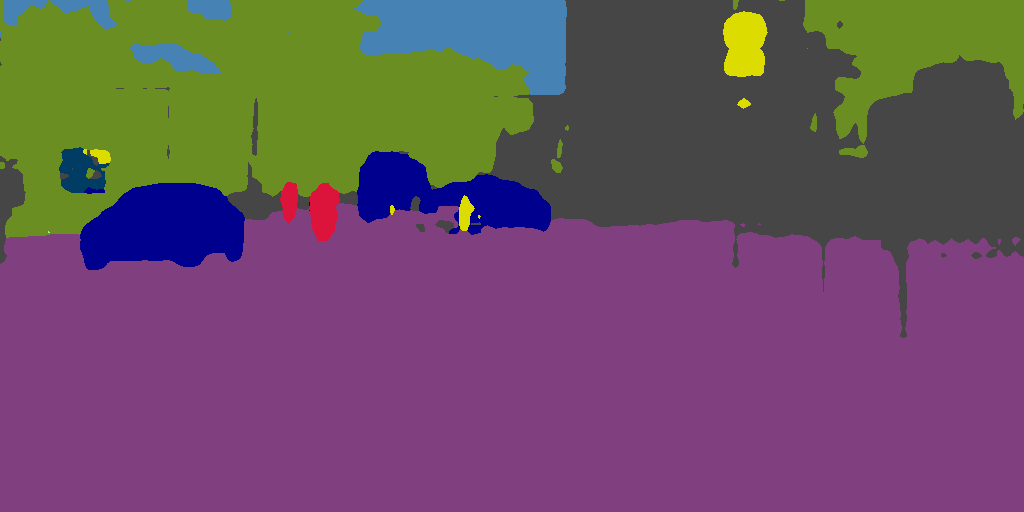}
    \end{subfigure}
    \hspace{0.01cm}
    \begin{subfigure}[b]{\textwidth}
        \includegraphics[width=\textwidth]{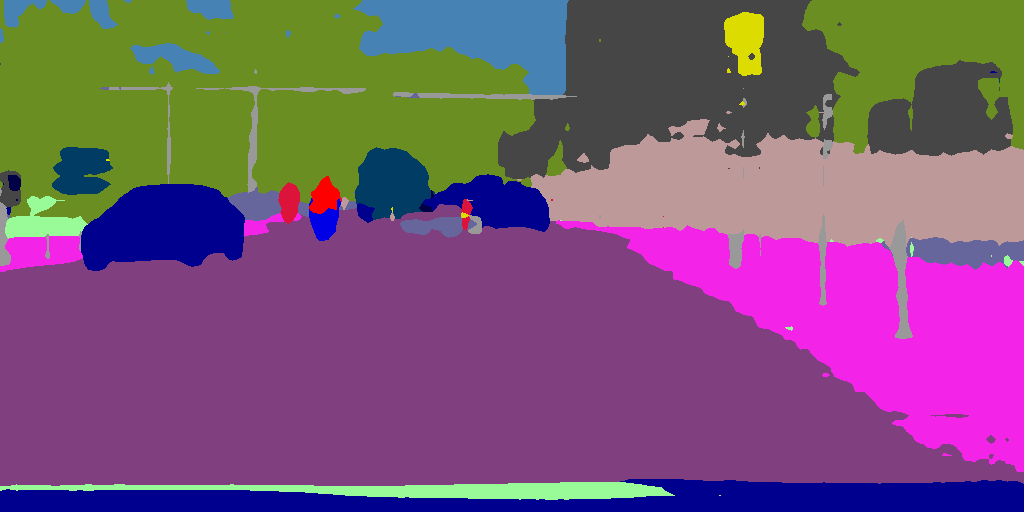}
    \end{subfigure}
    \end{adjustbox}
    
    \begin{adjustbox}{width=1\textwidth}
    \begin{subfigure}[b]{\textwidth}
        \includegraphics[width=\textwidth]{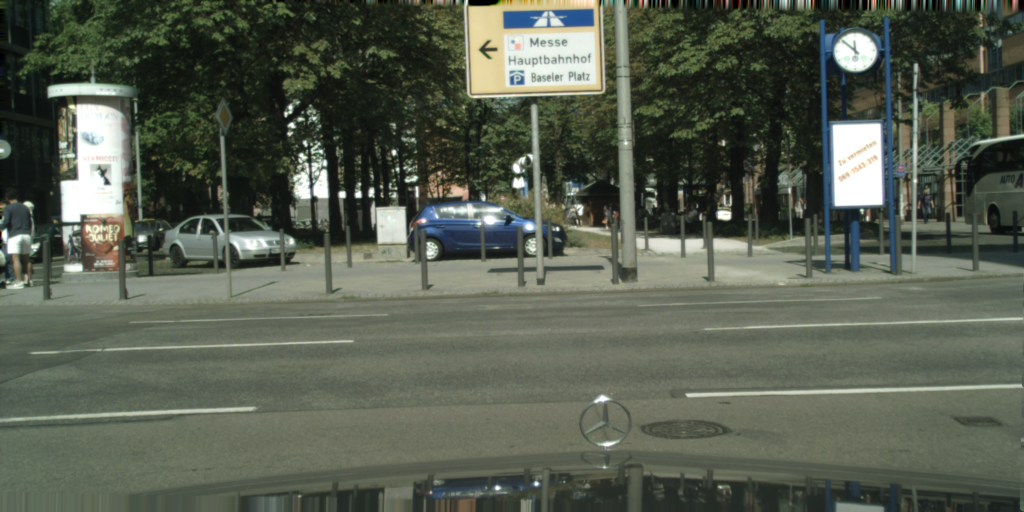}
    \end{subfigure}
    \hspace{0.01cm}
    \begin{subfigure}[b]{\textwidth}
        \includegraphics[width=\textwidth]{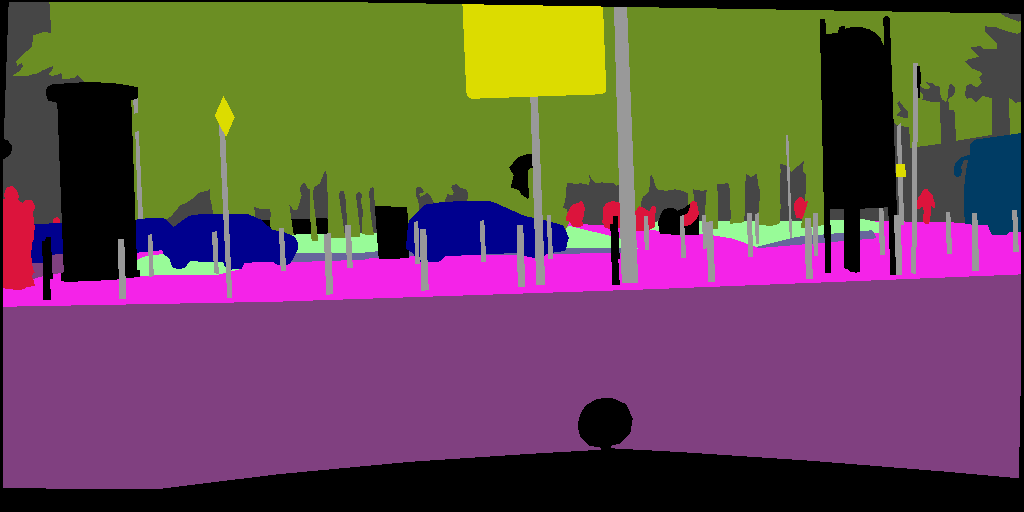}
    \end{subfigure}
    \hspace{0.01cm}
    \begin{subfigure}[b]{\textwidth}
        \includegraphics[width=\textwidth]{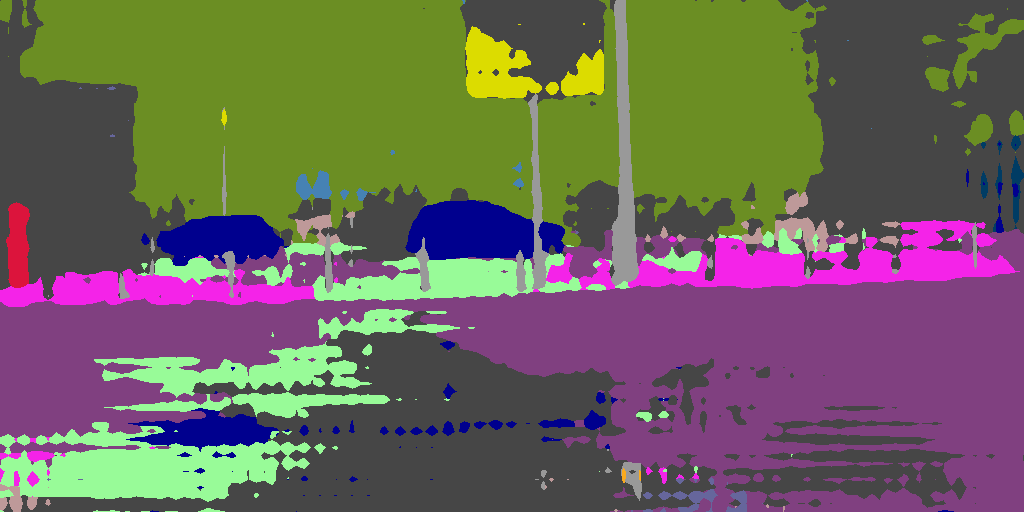}
    \end{subfigure}
    \hspace{0.01cm}
    \begin{subfigure}[b]{\textwidth}
        \includegraphics[width=\textwidth]{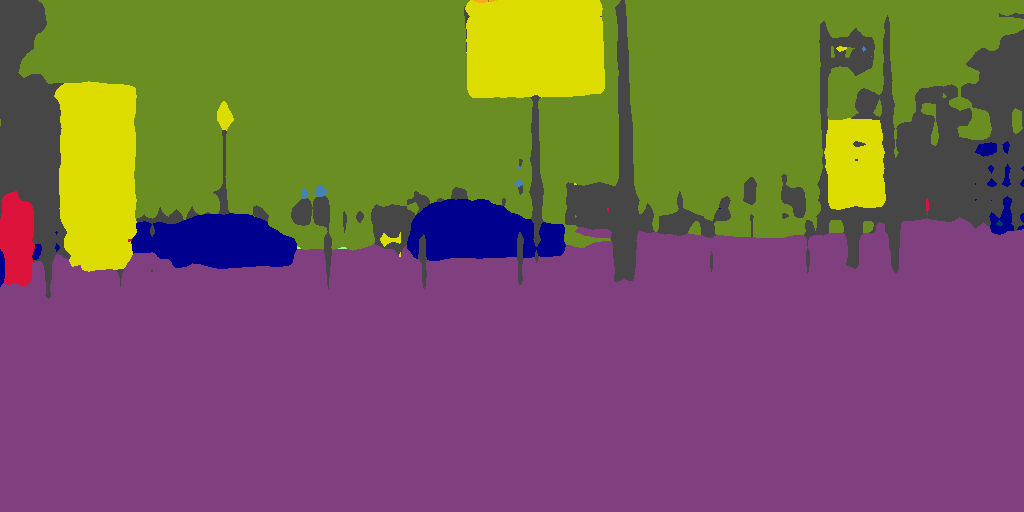}
    \end{subfigure}
    \hspace{0.01cm}
    \begin{subfigure}[b]{\textwidth}
        \includegraphics[width=\textwidth]{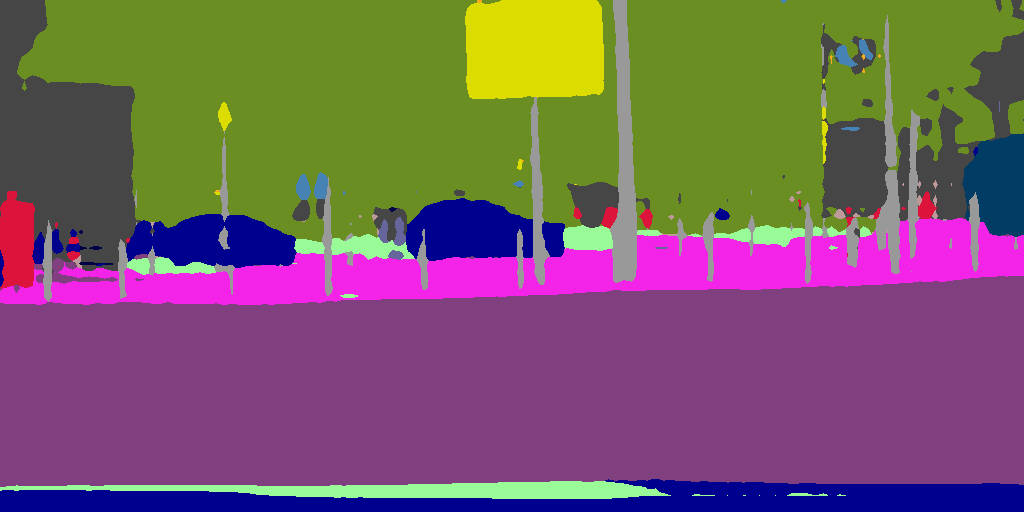}
    \end{subfigure}
    \end{adjustbox}

    \caption{Qualitative results of validation images from Cityscapes, when training models on the GTA5 dataset. The Naive Mixing model is consistently conflating some classes, sidewalks, for instance, are classified as road for all images. This issue is not shared with DACS.}
    \label{fig:DAresults}
\end{figure*}{}

Additionally, we also provide qualitative results for GTA5 $\rightarrow$ Cityscapes. Figure \ref{fig:DAresults} illustrates predictions on a few Cityscapes frames, made by DACS as well as by a source model (a model trained only supervised on the source domain). In the same figure, predictions made by a model trained using the Naive mixing described in Section \ref{sec:naivemix} are also presented. We can see that Naive Mixing performs better than the baseline, and DACS in turn outperforms Naive Mixing. This is also evident in Table \ref{tab:Ablation} in Section \ref{sec:ablation}, where results from Naive Mixing are shown. There, it is clear that some classes are conflated, for instance the 'sidewalk' class which obtains an IoU of almost 0\%, since it's almost always confused with 'road', which is also visible in Figure \ref{fig:DAresults}.

\subsection{SYNTHIA $\rightarrow$ Cityscapes Results}
For the SYNTHIA dataset, we also compare with the best reported performance of existing methods, regardless of network. Additionally, the SYNTHIA dataset only contains 16 of the 19 classes of Cityscapes, and some authors present results for these 16 classes while other present for only 13 of them. Because of these differences in benchmarking, we present results for both 13 and 16 classes for the DACS method. The results for the SYNTHIA dataset, computed with the same metrics (per-class IoU and mIoU) as the former dataset, are shown in Table \ref{tab:synthiaCSresults}, alongside results from several other existing works. 

When evaluating on all the 16 classes, DACS  obtainis an mIoU of 48.34\%, while also achieving the strongest per-class results for 4 out of the 16 classes. For the 13 class formulation, DACS obtains an mIoU of 54.81\%.

\begin{table*}[t]
    \centering
    \caption{SYNTHIA to Cityscapes, our results are averages from three runs, and shown as per-class IoU and mIoU for 13 and 16 classes. We also compare to several previous works.}
    \begin{adjustbox}{width=1\textwidth}
    \begin{tabular}{l|llllllllllllllll|l|l}
    \hline
                Method & Road & SW & Build & Wall\footnotemark & Fence\footnotemark[\value{footnote}] & Pole\footnotemark[\value{footnote}] & TL & TS & Veg & Sky  & Person & Rider & Car & Bus & MC & Bike & mIoU & mIoU \\ \hline
                
AdaptSegNet \cite{tsai2018learning}   & 84.3 & 42.7 & 77.5 & - & - &  - & 4.7 & 7.0 & 77.9 & 82.5 & 54.3 & 21.0 & 72.3 & 32.2 & 18.9 & 32.3 & 46.7 & -       \\ 
SIBAN \cite{luo2019significanceaware}   & 82.5 & 24.0 & 79.4 & - & - & - & 16.5 & 12.7 & 79.2 & 82.8 & 58.3 & 18.0 & 79.3 & 25.3 & 17.6 & 25.9 & 46.3 & -  \\ 
CLAN \cite{luo2018taking}   & 81.3 & 37.0 & 80.1 & - & - &  - & 16.1 & 13.7 & 78.2 & 81.5 & 53.4 & 21.2 & 73.0 & 32.9 & 22.6 & 30.7 & 47.8 & - \\ 
APODA \cite{yang2019adversarial}   & 86.4 & 41.3 & 79.3 & - & - & - & 22.6 & 17.3 & 80.3 & 81.6 & 56.9 & 21.0 & 84.1 & \textbf{49.1}  & 24.6 & 45.7 & 53.1 & - \\ 
PatchAlign \cite{tsai2019domain}   & 82.4 & 38.0 & 78.6 & 8.7 & 0.6 & 26.0 & 3.9 & 11.1 & 75.5 & 84.6 & 53.5 & 21.6 & 71.4 & 32.6 & 19.3 & 31.7 & 46.5 & 40.0 \\ 
AdvEnt \cite{vu2018advent}   & 85.6 & 42.2 & 79.7 & 8.7 & 0.4 & 25.9 & 5.4 & 8.1 & 80.4 & 84.1 & 57.9 & 23.8 & 73.3 & 36.4 & 14.2 & 33.0 & 48.0 & 41.2 \\ 
CBST \cite{zou2018domain}   & 68.0 & 29.9 & 76.3 & 10.8 & 1.4 & 33.9 & 22.8 & 29.5 & 77.6 & 78.3 & 60.6 & 28.3 & 81.6 & 23.5 & 18.8 & 39.8 & 48.9 & 42.6 \\ 
MRKLD \cite{zou2019confidence}   & 67.7 & 32.2 & 73.9 & 10.7 & 1.6 & \textbf{37.4} & 22.2 & 31.2 & 80.8 & 80.5 & 60.8 & 29.1 & 82.8 & 25.0 & 19.4 & 45.3 & 50.1 & 43.8 \\ 

CADASS \cite{yang2020contextaware}   & 82.5 & 42.2 & 81.3 & - & - & - & 18.3 & 15.9 & 80.6 & 83.5 & 61.4  & 33.2 & 72.9 & 39.3 & 26.6 & 43.9 & 52.4 & -\\ 

MRNet \cite{zheng2019unsupervised}   & 82.0 & 36.5 & 80.4 & 4.2 & 0.4 & 33.7 & 18.0 & 13.4 & 81.1 & 80.8 & 61.3 & 21.7 & 84.4 & 32.4 & 14.8 & 45.7 & 50.2 & 43.2 \\ 
R-MRNet \cite{zheng2020rectifying}   & 87.6 & 41.9 & 83.1 & 14.7 & 1.7 & 36.2 & \textbf{31.3} & 19.9 & 81.6 & 80.6 & 63.0 & 21.8 & 86.2 & 40.7 & 23.6 & \textbf{53.1} & 54.9 & 47.9 \\ 

PIT \cite{lv2020PIT} & 83.1 & 27.6 & 81.5 & 8.9  & 0.3 & 21.8 & 26.4 & \textbf{33.8} & 76.4 & 78.8 & 64.2 & 27.6 & 79.6 & 31.2 & 31.0 & 31.3 & 51.8 & 44.0 \\ 

SIM \cite{wang2020differential} & 83.0 & 44.0 & 80.3 & - & - & - & 17.1 & 15.8 & 80.5 & 81.8 & 59.9 & 33.1 & 70.2 & 37.3 & 28.5 & 45.8 & 52.1 & - \\ 

FDA \cite{yang2020fda} & 79.3 & 35.0 & 73.2 & - & - & - & 19.9 & 24.0 & 61.7 & 82.6 & 61.4 & 31.1 & 83.9 & 40.8 & \textbf{38.4} & 51.1 & 52.5 & - \\

Yang et al. \cite{yang2020labelddriven} & 85.1 & 44.5 & 81.0 & - & - & - & 16.4 & 15.2 & 80.1 & 84.8 & 59.4 & 31.9 & 73.2 & 41.0 & 32.6 & 44.7 & 53.1 & - \\ 

Kim et al. \cite{DBLP:conf/cvpr/KimB20a} & \textbf{92.6} & \textbf{53.2} & 79.2 & - & - & - & 1.6 & 7.5 & 78.6 & 84.4 & 52.6 & 20.0 & 82.1 & 34.8 & 14.6 & 39.4 & 49.3 & - \\

CAG-UDA \cite{DBLP:conf/nips/ZhangZ0T19} & 84.7 & 40.8 & 81.7 & 7.8 & 0.0 & 35.1 & 13.3 & 22.7 & \textbf{84.5} & 77.6 & 64.2 & 27.8 & 80.9 & 19.7 & 22.7 & 48.3 & - & 44.5 
\\ 

IAST \cite{DBLP:journals/corr/abs-2008-12197} & 81.9 & 41.5 & \textbf{83.3} & 17.7 & \textbf{4.6} & 32.3 & 30.9 & 28.8 & 83.4 & 85.0 & 65.5 & 30.8 & \textbf{86.5} & 38.2 & 33.1 & 52.7 & \textbf{57.0} & \textbf{49.8}  \\
\hline

Source & 36.30 & 14.64 & 68.78 & 9.17 & 0.20 & 24.39 & 5.59 & 9.05 & 68.96 & 79.38 & 52.45 & 11.34 & 49.77 & 9.53 & 11.03 & 20.66 & 33.65 & 29.45 \\

DACS & 80.56 & 25.12 & 81.90 & \textbf{21.46} & 2.85 & 37.20 & 22.67 & 23.99 & 83.69 & \textbf{90.77} & \textbf{67.61} & \textbf{38.33} & 82.92 & 38.90 & 28.49 & 47.58 & 54.81 & 48.34 \\
\hline

    \end{tabular}
    \end{adjustbox}
    \label{tab:synthiaCSresults}
\end{table*}{}

\subsection{Evaluation Conditions}

Many existing methods in Tables \ref{tab:GTACSresults} and \ref{tab:synthiaCSresults}, report their results from when using early stopping based on the same validation set that is used for the final evaluation (there is no publicly available test set for Cityscapes). We believe that this is not a fair evaluation, as a high performance on the validation set does not necessarily mean high performance on all data, but could just mean that the model is performing well on those exact images. Using early stopping in our case would increase the results substantially: for GTA5 it would increase to 35.68\% for the baseline just trained on the source data and to 53.84\% for the DACS results. For SYNTHIA, the source baseline would increase to 32.85\%, and the results for 13 classes would increase to 55.98\% and for 16 classes to 49.10\%. The reason for the considerable performance increase from early stopping is that performance on the validation set fluctuates a lot over the course of training, rather than that the model is overfitting the training data.

Similarly, many existing methods in UDA report final results based on the top-performing model from sensitive hyper-parameter tuning on the aforementioned validation set. Reporting performance obtained after hyper-parameter tuning would skew results in the same way as reporting the highest among multiple experiments rather than averages, as one would then exploit the variance between 
hyper-parameters. DACS was evaluated as an average of three runs, with a top performing model of 54.09\% for GTA5, 55.21\% for Synthia with 13 classes, and 49.07\% for Synthia with 16 classes. 


\begin{table*}
    \centering
    \caption{Results from experiments when using Naive Mixing, only using pseudo-labelling and when applying CutMix or CowMix instead of the default ClassMix. All experiments are performed on GTA5 $\rightarrow$ Cityscapes. For comparison our results from Table \ref{tab:GTACSresults} are also included.}
    \begin{adjustbox}{width=1\textwidth}
    \begin{tabular}{l|lllllllllllllllllll|l}
    \hline
                Method & Road & SW & Build & Wall & Fence & Pole & TL & TS & Veg & Terrain & Sky  & Person & Rider & Car  & Truck & Bus  & Train & MC & Bike & mIoU \\ \hline

Source   & 63.31 & 15.65 & 59.39 & 8.56 & 15.17 & 18.31 & 26.94 & 15.00 & 80.46 & 15.25 & 72.97 & 51.04 & 17.67 & 59.68 & 28.19 & 33.07 & \textbf{3.53} & 23.21 & 16.73 & 32.85 \\

Naive Mixing& 84.78 & 0.00 & 82.81 & 0.34 & 0.05 & 10.56 & 47.96 & \textbf{58.86} & 86.87 & 8.08 & 90.99 & 56.09 & 0.00 & 86.92 & 40.45 & 11.38 & 0.00 & 0.45 & 0.00 & 35.08\\

Pseudo-labelling   & 85.14 & 0.03 & 75.84 & 0.35 & 0.03 & 0.23 & 3.19 & 1.30 & 78.05 & 36.53 & 65.78 & 4.89 & 0.01 & 79.17 & 4.24 & 1.30 & 0.00 & 0.26 & 0.02 & 22.97 \\

DACS + CutMix & 86.29 & 36.92 & 85.47 & \textbf{32.63} & 31.95 & 27.86 & 44.17 & 30.19 & 80.34 & 25.28 & 85.79 & 65.03 & 34.99 & \textbf{90.08} & 46.69 & \textbf{50.40} & 3.48 & \textbf{36.90} & 30.62 & 48.69 \\

DACS + CowMix &  \textbf{90.34} & 33.93 & 84.12 & 27.37 & 36.33 & 31.47 & 44.15 & 28.03 & 85.56 & 38.94 & 87.94 & 63.82 & 24.77 & 89.48 & \textbf{47.41} &  49.54 & 0.16 & 36.78 & 17.66 & 48.30 \\

Distribution alignment& 85.05 & 28.88 & 86.79 & 16.92 & 36.89 & 30.38 & \textbf{49.73} & 53.91 & 85.61 & 32.20 & \textbf{92.78} & 66.61 & 23.53 & 84.00 & 34.81 & 27.70 & 0.20 & 16.65 & \textbf{60.08} & 48.04 \\

DACS& 89.90 & \textbf{39.66} & \textbf{87.87} & 30.71 & \textbf{39.52} & \textbf{38.52} & 46.43 & 52.79 & \textbf{87.98} & \textbf{43.96} & 88.76 & \textbf{67.20} & \textbf{35.78} & 84.45 & 45.73 & 50.19 & 0.00 & 27.25 & 33.96 & \textbf{52.14} \\
\hline

    \end{tabular}
    \end{adjustbox}
    \label{tab:Ablation}
\end{table*}{}

\footnotetext[1]{Excluded when calculating mIoU for 13 classes.}

\subsection{Similarity Between Source and Target}
We note that our results increase more in relation to previous works for GTA5 $\rightarrow$ Cityscapes than it does for SYNTHIA $\rightarrow$ Cityscapes. We hypothesise that this depends on the similarity between the source and target data, namely that GTA5 images are more similar to Cityscapes than SYNTHIA images are. This is clearly helpful in UDA and particularly so when mixing images between domains, since the mixed images will be more sensible if objects end up in locations that are reasonable. This can be quantified by the spatial distribution of classes: where in the images certain classes appear. Cityscapes and GTA5 are very similar in that road is always down, sky is always up and cars are always vertically near the center of the images. This is less the case for SYNTHIA, however, as the images are taken from different perspectives, including from the ground and from above. Hence, when pasting objects from SYNTHIA to Cityscapes, it is likely that the resulting images will be nonsensical, which we believe is detrimental for training. This means that the performance of DACS might not be as high for tasks with larger domain shifts. We leave the investigation of this for future work.

\subsection{Additional Experiments}
\label{sec:ablation}
In order to further evaluate DACS, and to better understand the source of the class conflation problem, additional experiments are performed and presented in Table \ref{tab:Ablation}. Results are shown for the Naive Mixing explained in Section \ref{sec:naivemix} as well as for using only pseudo-labelling without any mixing and for using different mixing strategies.

As can be seen in Table \ref{tab:Ablation}, the performance is significantly stronger for DACS than it is for Naive Mixing. As stated in Section \ref{sec:naivemix}, the most important reason for this is that Naive Mixing conflates several of the classes, which impacts the overall performance considerably. This is clear from the per-class IoUs in the table, where seven of the classes have scores below 1\% for Naive Mixing. In Figure \ref{fig:DAresults}, we can clearly see an illustration of the class conflation problem, where for example the 'sidewalk' class is not predicted by Naive Mixing in any of the frames, instead being classified as road. In contrast, DACS is able to correctly discern between these classes in all frames.

Since Naive Mixing is mixing images based on predictions, we investigate if the problem of class conflation could be related to, or made worse by, the mixing component. 
We observe that when just using pseudo-labelling (that is removing the mixing component), even more classes stop being predicted by the network, with overall performance becoming worse than the source baseline. Therefore, it is reasonable to assume that it is the pseudo-labelling component, and not the mixing, that cause class conflation, similar to the conclusion made in \cite{zou2018domain}. It is possible that incorporating existing techniques for pseudo-labelling would solve this issue. Though as previously stated, our proposed method of mixing images across domains offers a simple correction purely by changing the nature of the augmentation technique.

A different way to solve the problem of conflating classes would be to impose a prior class distribution on the pseudo-labels. This has been done previously for the same task in the context of entropy regularization \cite{vu2018advent}. It is therefore insightful to investigate if a similar solution can correct pseudo-labels in Naive Mixing. To this end we use Distribution Alignment, as used in \cite{ReMixMatch}, meaning that a distribution of classes is forced upon the predictions. This is a different way of injecting entropy into the pseudo-labels, meaning that it could also help the network avoid class conflation.
We perform experiments where the ground-truth distribution $p$ of the target dataset is used to guide the training. This is done for each sample by transforming the output prediction $q$ for each pixel into $\tilde{q} = \text{Normalize}(q \times p/\tilde{p})$, where $\tilde{p}$ is a running average of all predictions made by the network on the target data. 
The setup is otherwise identical to when using Naive Mixing. The results from this are shown in Table \ref{tab:Ablation}. In our case, this is not a legitimate method, as the ground-truth class distribution would not be known for an unlabelled dataset in a realistic setting. However, it is interesting to see that this approach also solves the issue of conflating classes, as all classes are represented in the results. This further strengthens our hypothesis that artificial injection of entropy in training can help the network avoid class conflation. 

When replacing ClassMix \cite{ClassMix} with other mixing techniques such as CutMix \cite{CutMix}, or CowMix \cite{MilkingCowMask}, performance drops slightly but the class conflation issue remains solved independent of mixing strategy, as can be seen in Table \ref{tab:Ablation}. This both illustrates the general applicability of cross domain mixing, and that the choice of mixing strategy is relevant for performance. Incorporating other mixing strategies with DACS has the potential for even stronger performance, and presents an interesting direction for future research.

\section{Conclusion}
We have proposed DACS, Domain Adaptation via Cross-domain mixed Sampling, a novel algorithm for unsupervised domain adaptation in semantic segmentation. We show how a naive adaptation of mixing-based consistency regularization as used in SSL to UDA results in systematic problems in the predictions, and detail the changes performed in order to correct these issues. Furthermore, we perform an evaluation of DACS for two popular domain adaptation benchmarks, GTA5 $\rightarrow$ Cityscapes, where it outperforms existing methods and pushes the state of the art, and SYNTHIA $\rightarrow$ Cityscapes.

\vspace{0.4cm}
\noindent {\bf Acknowledgements.} This work was partially supported by the Wallenberg AI, Autonomous Systems and Software Program (WASP) funded by the Knut and Alice Wallenberg Foundation.

{\small
\bibliographystyle{ieee_fullname}
\bibliography{egbib}
}

\end{document}